\documentclass[letterpaper, 10 pt, conference]{ieeeconf}
\IEEEoverridecommandlockouts                              

\usepackage{times}
\usepackage{url}
\usepackage[pagebackref=true,breaklinks=true,letterpaper=true,colorlinks,bookmarks=false]{hyperref}
 \usepackage{epsfig}
 \usepackage{epstopdf}
\usepackage{graphicx}
\usepackage{amsmath}
\usepackage{amssymb}
\usepackage{ mathrsfs }
\usepackage{color}
\usepackage{algpseudocode}
\usepackage{algorithm}
\usepackage{placeins}
\usepackage{rotating}
\usepackage{capt-of,etoolbox}
\usepackage{threeparttable}
\usepackage{booktabs}
\usepackage{mathtools}
\usepackage{amsfonts} 
\usepackage{siunitx}
\usepackage{caption}
\newcommand\norm[1]{\left\lVert#1\right\rVert}
\DeclareSIUnit\Molar{\textsc{m}} 
\DeclareSIUnit{\pH}{pH}
\DeclareSIUnit{\pixel}{px}

\newcommand{\G}{\mathcal{G}}
\newcommand{\V}{\mathcal{V}}
\newcommand{\E}{\mathcal{E}}
\newcommand{\W}{\mathcal{W}}
\newcommand{\Rbb}{\mathbb{R}}

\title{\LARGE \bf
Graph Based Sinogram Denoising for Tomographic Reconstructions
}

\title{\LARGE \bf
Graph Based Sinogram Denoising for Tomographic Reconstructions
}

\author{Faisal Mahmood$^{*1}$, Nauman Shahid$^{*2}$, Pierre Vandergheynst$^{\dagger2}$ and Ulf Skoglund$^{\dagger1}$
\thanks{*Contributed Equally $\dagger$Co-senior Author}
\thanks{F.M. and U.S. were supported by Japanese Government OIST Subsidy for Operations(Skoglund U.) under grant number 5020S7010020. F.M. was additionally supported by the OIST PhD Fellowship. N.S. was supported by SNF grant 200021\_154350/1 for the project ``Towards signal processing on graphs''}
\thanks{$^{1}$Structural Cellular Biology Unit, Okinawa Institute of Science and Technology (OIST), 1919-1 Tancha, Onna, Okinawa 904-0495, Japan.
        {\textit{faisal.mahmood@oist.jp, ulf.skoglund@oist.jp}}}%
\thanks{$^{2}$Signal Processing Laboratory 2 (LTS2), EPFL STI IEL, Lausanne, CH-1015, Switzerland.
        {\textit{nauman.shahid@epfl.ch, pierre.vandergheynst@epfl.ch}}}%
}

\begin{document}

\maketitle
\thispagestyle{empty}
\pagestyle{empty}

\begin{abstract}

Limited data and low dose constraints are common problems in a variety of tomographic reconstruction paradigms which lead to noisy and incomplete data. Over the past few years sinogram denoising has become an essential pre-processing step for low dose Computed Tomographic (CT) reconstructions. We propose a novel sinogram denoising algorithm inspired by the modern field of signal processing on graphs. Graph based methods often perform better than standard filtering operations since they can exploit the signal structure. This makes the sinogram an ideal candidate for graph based denoising since it generally has a piecewise smooth structure. We test our method with a variety of phantoms and different reconstruction methods. Our numerical study shows that the proposed algorithm improves the performance of analytical filtered back-projection (FBP) and iterative methods ART (Kaczmarz) and SIRT (Cimmino). We observed that graph denoised sinogram always minimizes the error measure and improves the accuracy of the solution as compared to regular reconstructions.

\end{abstract}


%
\IEEEpeerreviewmaketitle

\section{Introduction}
Low dose, limited and incomplete data are common problems in a variety of tomographic reconstruction paradigms. Computerized Tomography (CT) and Electron Tomography (ET) reconstructions are usually ill-posed inverse problems and encounter significant amounts of noise \cite{hsieh_computed_2009}\cite{fernandez_computational_2012}. During data collection individual projections are usually marred by noise due to low illumination, which stems from low electron or x-ray dose. Low dose CT projections  have a low signal to noise ratio (SNR) and render the reconstruction erroneous and noisy. Traditional scanners usually employ an analytical filtered back-projection (FBP) based reconstruction approach which performs poorly in limited data and low SNR situations \cite{natterer_mathematics_1986}. Over the past few years there has been a significant effort to reduce the dose for CT reconstructions because of associated health issues \cite{berrington_de_gonzalez_projected_2009}. Current efforts to reduce the dose for CT reconstructions can be divided into three categories \cite{karimi_sinogram_2016}\cite{fessler_statistical_2000}: 1) Pre-processing based methods which tend to improve the raw data (sinogram) followed by standard FBP based reconstruction \cite{wang_&lt;title&gt;sinogram_2005}\cite{bjorck_accelerated_1979}. 2) Denoising of reconstructed tomograms in image domain. 3) Iterative image reconstruction and statistical methods\cite{beister_iterative_2012}. Usually, a combination of the methods enlisted above render the best results. 
Denosing of the sinogram has been proposed in several studies using a variety of extensively studied approaches. These algorithms vary from simple adaptive filtering and shift-invariant low-pass filters to computationally complex bayesian methods \cite{la_riviere_penalized-likelihood_2005}\cite{la_riviere_penalized-likelihood_2006}. Other approaches involve Fourier and wavelet transform based multi-resolution methods \cite{natterer_mathematics_1986} and denoising projection data in Radon space\cite{wang_experimental_2008}.

The emerging field of signal processing on graphs \cite{shuman_emerging_2012} has made it possible to solve the inverse problems efficiently by exploiting the hidden structure in the data. One such problem is the graph based denoising which tends to perform better than the simple filtering operations due to its inherent nature to exploit the signal structure \cite{shuman_emerging_2012}. In this study we propose a new method to remove noise from the raw data (sinogram) using graph based denosing. Our proposed denoising is general in the sense that it can be applied to the raw data independent of the type of data or application under consideration. Furthermore, as shown in the experiments, it is independent of the reconstruction method. In fact, our denoising method tends to improve the quality of several standard tomographic reconstruction algorithms.

\section{Proposed Method}
 Let $S \in \Re^{p \times q}$ be the sinogram corresponding to the projections of the sample $x \in \Re^{n \times n}$ being imaged,  where $p$ is the number of rays passing through $x$ and $q$ is the number of angular variations at which $x$ has been imaged.   Let $b \in \Re^{pq}$ be the vectorized measurements or projections and $A \in \Re^{pq \times n^2}$ be the sparse projection operator. Then, the goal in a typical CT or ET based reconstruction method is to recover the sample $x$ from the projections $b$. Of course, one needs to solve a highly under-determined inverse problem for this type of reconstruction, which can be even more challenging if the projections $b$ are noisy. To circumvent the problem of noisy projections, we propose a two-step methodology for the reconstruction.  \textit{1) Denoise the sinogram $S$ using graph total variation regularization. 2) Reconstruct the sample $x$ from denoised projections using any standard reconstruction method.}
 
\begin{figure}[t!]
\centering
\includegraphics[width=0.48\textwidth]{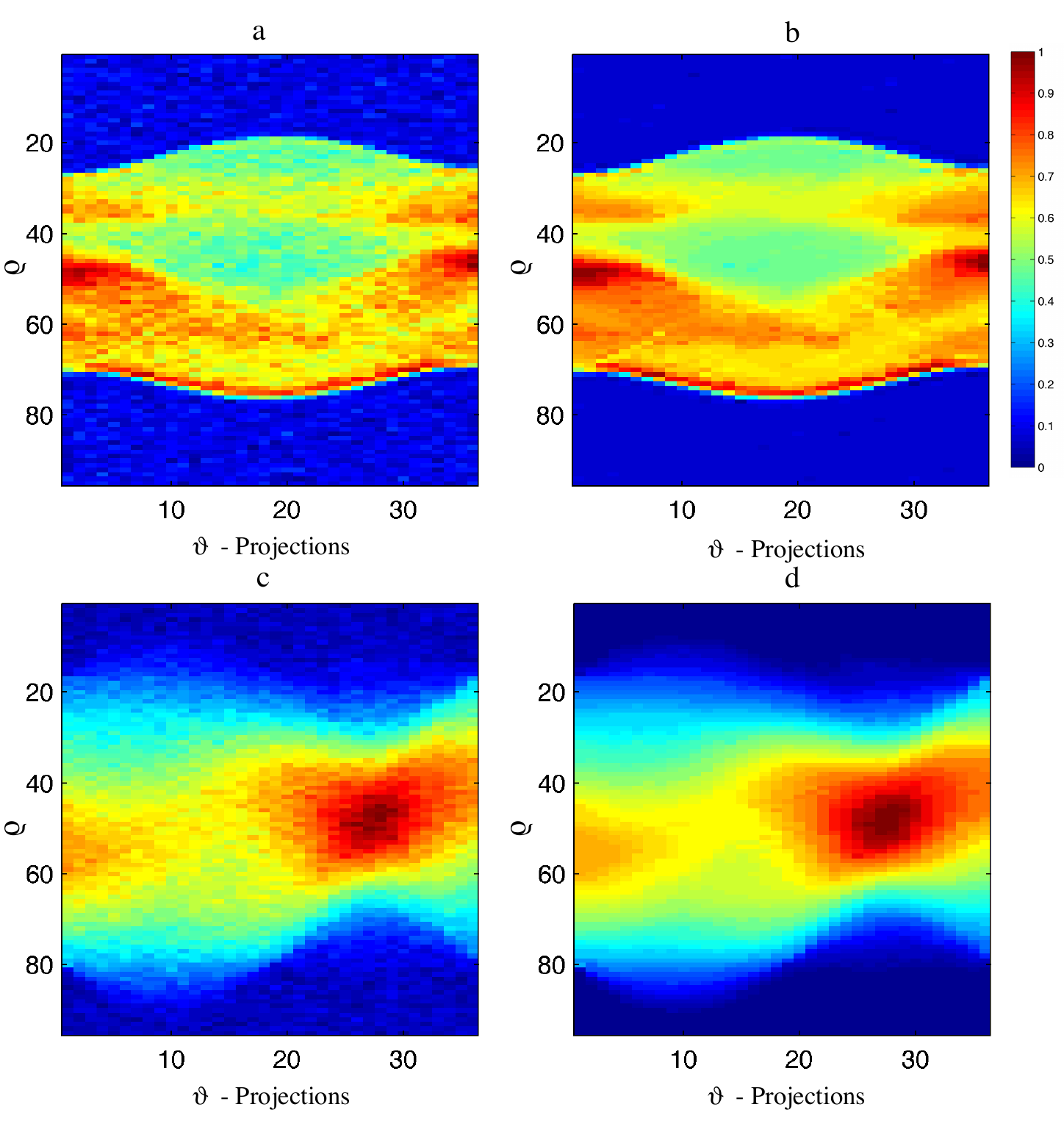}
\caption{The first row shows the noisy (relative noise 0.08) and denoised sinograms for the Shepp-Logan phantom and the second row for smooth phantom. Clearly, the projections have a smooth structure which when exploited denoises these sinograms significantly.}
\label{fig:projections}
\end{figure}

 While, we explain our proposed method in detail in the following section, we motivate the denoising method here via Fig. \ref{fig:projections} which shows sinograms corresponding to the Modified Shepp-Logan and Smooth phantoms.  It can be observed that the sinograms have a piecewise smooth structure. If a sinogram is treated as an image, it can be said that some patches of the sinogram are similar to other patches for a given phantom. This structure can be exploited in the form of  a pairwise similarity graph constructed between the patches of the sinogram and then used to denoise it via \textit{graph regularization}.
 
\section{Graph Based Sinogram Denoising}
\subsection{A Brief Introduction to Graphs}\label{sec:graph_notation}
A graph is a tupple $\mathcal{G}=\{ \V,\E,\mathcal{W}\}$ where $\V$ is a set of vertices, $\E$ a set of edges,
and $\W : \V \times \V \rightarrow \Rbb_+$ a weight function. The vertices are indexed from $1,\dots, |\V|$ and each entry  of the weight matrix $W \in \mathbb R^{|\V|\times |\V|}_+$ contains the weight of the edge connecting the corresponding vertices: $W_{i,j} = \W (v_i,v_j)$. If there is no edge between two vertices, the weight is set to $0$. We assume $W$ is symmetric, non-negative and with zero diagonal. We denote by $i\leftrightarrow j$ that node $v_i$ is connected to node $v_j$.  For a vertex $v_i\in \V$, the degree $d(i)$ is defined as the sum of the weights of incident edges: $d(i)=\sum_{j \leftrightarrow i } W_{i,j}$. 
A graph signal is defined as a function $f: \V \rightarrow  \Rbb$ assigning a value to each vertex. It is convenient to consider a signal $f$ as a vector of size $|\V|$ with the $i^{\mathrm{th}}$ component representing the signal value at the $i^{\mathrm{th}}$ vertex. 


\begin{algorithm}
\captionsetup{font=small,skip=0pt}
\caption{Graph Total Variation Denoising}
\label{CHalgorithm}
\begin{algorithmic}
\State INPUT: $u_0 = 0$, $\epsilon > 0$, OUTPUT: $x_j$
\For{ $j = 0,\dots J-1$ }
\State $x_{j} =  Y - \nabla^{*}_{\G}(u_j) $
\State $r_j = u_j\tau + \nabla_{\G}(x_j)$
\State $s_{j} = \max (r_j - \gamma\tau, 0) $
\State $u_{j+1} = \frac{1}{\tau}(r_j - s_j)$
\State $F_{j + 1} = \|b-x_j\|^{2}_2 + \gamma \|u_{j+1}\|_1$
\If{$\frac{\|F_{j+1} - F_{j}\|_F^2}{\| F_{j}\|_F^2}<\epsilon$}
\State BREAK\textbf{}
\EndIf
\EndFor
\end{algorithmic}
\end{algorithm}

\subsection{Graph Construction}\label{sec:graphs}
For our purpose the  graph $\G$ is a patch graph, i.e, a graph between the patches of sinogram $S$ and it is built using a three-step  strategy. In the first step the sinogram $S \in \Re^{p\times q}$ is divide into $pq$ overlapping patches. Let $s_i$ be the patch of size $l \times l$ centered at the $i^{th}$ pixel of $S$ and assume that all the patches are vectorized, i.e, $s_i \in \Re^{l^2}$. In the second step the search for  the closest neighbours  for all the vectorized patches is performed using the Euclidean distance metric.  Each  $s_i$ is connected to its $K$ nearest neighbors $s_j$, resulting in $|\mathcal{E}|$ number of connections. In the third step the graph weight matrix $W$ is computed as
\begin{equation*}
W_{i,j} = \begin{cases}
\exp\Big(-\frac{ \|(s_i-s_j)\|^{2}_{2}}{ \sigma^{2}}\Big) & \text{if $s_j$ is connected to $s_i$}\\
0 & \text{otherwise.}\\
\end{cases}
\end{equation*}
The parameter $\sigma$ can be set experimentally as the average distance of the connected samples. This procedure has a complexity of $\mathcal{O}(pq e)$ and each $W_{i,j}$ can be computed in parallel.

\subsection{Graph Total Variation Denoising}

Once the graph $\G$ has been constructed, we solve the following optimization problem to denoise $b$.

\begin{align}\label{eq:gtv}
& \min_{z} \|z - b\|^2_2 + \gamma \|\nabla_{\G} z\|_1,
\end{align} 
which can also be written as,
\begin{align}
& \min_{z} \|z - b\|^2_2 + \gamma \sum_i \sum_j \sqrt{W_{i,j}}|z_i - z_j|,
\end{align} 

where $\nabla_{\G}z$ denotes the graph total variation of $z$. The parameter $\gamma$ controls the  amount of smoothing on $z$. Higher levels of noise require higher $\gamma$.

In simple words, as $\ell_1$ norm promotes sparsity, problem  \eqref{eq:gtv} tends to smooth the projections $b$, such that the new projections $z$ have sparse graph gradients. Intuitively, this makes sense, as the sinograms (Fig. \ref{fig:projections}) have a piecewise smooth structure. Thus, all the strongly connected patches in $S$ will have a similar structure in $z$ (zero graph gradients) whereas the weakly connected patches in $S$ will have different structure in $z$ (non-zero graph gradients).  

The solution to problem \eqref{eq:gtv} is given by the algorithm given below. The algorithm requires $\nabla^{*}_{\G}(\cdot)$, which is the adjoint operator (divergence) of $\nabla_{\G}$  and $\tau$ is the spectral norm (maximum eigenvalue) of $\nabla_{\G}$, i.e, $\tau = \|\nabla_{\G}\|_2$. The algorithm is a simple application of the proximal splitting methods commonly used in signal processing \cite{combettes_proximal_2011}. Note that the solution to the $\ell_1$ norm is a simple element-wise soft-thresholding operation. 

\begin{figure*}
  \includegraphics[width=\textwidth]{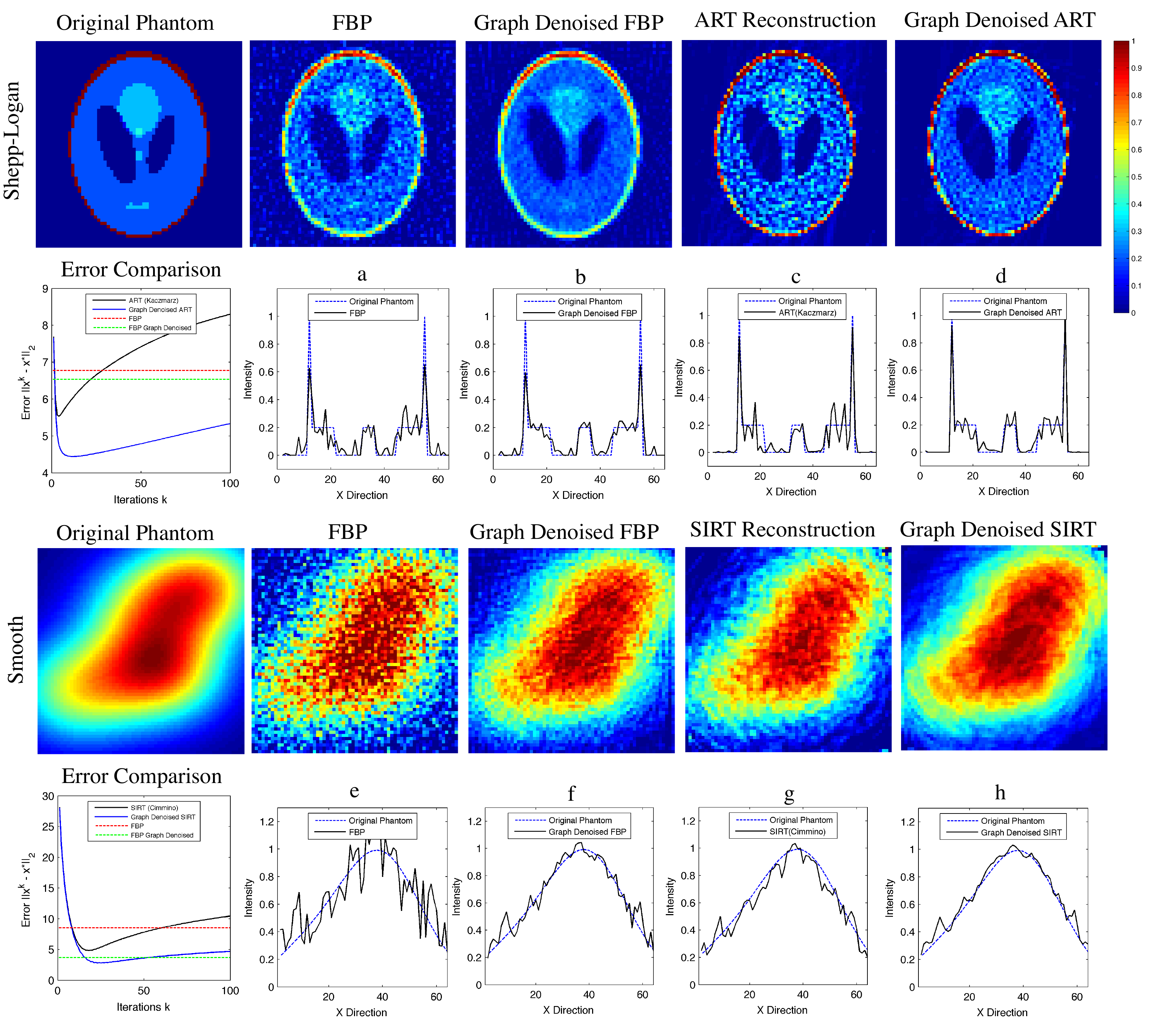}
  \caption{$\ell_2$ reconstruction error variations with the iterations $k$, the recovered phantoms and their intensity profiles  for the following methods: 1) Filtered back projection (FBP) 2) Graph denoised FBP (FBP-GD) 3) ART / SIRT reconstruction (without denosing) and 4) Graph denoised  ART / SIRT reconstruction (ART-GD, SIRT-GD). The images can be best viewed in colour on the electronic version of the paper.
}
  \label{fig:results}
  \end{figure*}

\section{Reconstruction Phase}
We reconstruct the refined sinogram using different analytical and iterative methods. Analytical methods such as FBP \cite{brooks_theory_1975} are commonly used in CT scanners and other tomographic modalities since they are simple and computationally trivial. However, such methods give poor reconstructions when data is limited and noisy. In such cases iterative methods have to be used. Iterative row action methods such as algebraic reconstruction technique (ART) \cite{gordon_algebraic_1970} and simultaneous reconstruction technique (SIRT) \cite{gilbert_iterative_1972} are commonly used due to their semi-convergance and regularization capabilities. ART or Kaczmarz method operates on each row/equation of the linear system defined in section I and treats it as a hyperplane in vector space. Starting from an initial guess each iteration entails sweeping all rows by taking orthogonal projections successively until it converges to a solution, this can be defined as follows:

\begin{equation}\label{eq:ART}
  \begin{aligned}
  X \leftarrow  X + \lambda {\frac{b_i-a{_i^T} x}{\norm{a_i}^2_2}} a_i \qquad \lambda \in (0,2)
  \end{aligned}
\end{equation}

The fact that during the early iterations Kaczmarz quickly semi-converges to an approximate solution makes it an ideal method to test sinogram denoising. A refined sinogram should semi-converge to a lower error quicker and diverge from the most accurate solution slower. The class of SIRT methods access all rows simultaneously and generally have better regularization capabilities as compared to ART. Although there are several different variants of SIRT we used Cimmino's method \cite{cimmino_calcolo_1938} which can be defined as follows:

\begin{equation}
  \begin{aligned}
  x^{k+1} =  x^{k} + \lambda_{k} \frac{1}{m} {\sum_{n=1}^{m} w_{i}\frac{b_i-\big \langle a{_i}, x^k \big \rangle}{\norm{a_i}^2_2}} a_i .
  \end{aligned}
\end{equation}

Cimmino's method usually converges faster than ART and has similar semi-convergance properties.

\section{Experimental Results and Conclusions}
We perform our denoising and reconstruction experiments on 5 different types of phantoms 1) Shepp-Logan \cite{shepp_fourier_1974} 2) smooth 3) binary 4) grains and 5) fourphases \cite{hansen_air_2012} each of the size $64 \times 64$. The sinograms $S$ are of size $95 \times 36$, where each column corresponds to the projections at each of the equally spaced 36 angles from 0 to 180 degrees. Random Gaussian noise with mean 0 and variance adjusted to the $5\%$ and $8\%$ of the norm of $S$ is added into $S$. For the graph based total variation denoising stage of our experiments, the sinogram is divided into $95 \times 36 = 3420$ overlapping patches of size $3\times 3$. The graph $\G$ is constructed between the $3420$ patches with 10 nearest neighbors ($K = 10$) and $\sigma$ for the weight matrix (Section \ref{sec:graphs}) is set to the average distance of the 10-nearest neighbors. Various values of $\gamma$ in the range of $[0,10]$ are tested for denoising (Algorithm 1). The best $\gamma$ is selected based on the minimum $\ell_2$ reconstruction for the phantoms. Fig. \ref{fig:projections} shows the quality of graph total variation based denosing for the Modified Shepp-Logan sinogram (row 1) and smooth phantom sinogram (row 2). The relative noise for each of the two sinograms is $0.08$. 

Table \ref{tab:test2} and Fig. \ref{fig:results} present $\ell_2$ reconstruction error variations with the iterations $k$, the recovered phantoms and their intensity profiles  for the following methods: 1) Filtered back projection (FBP) 2) Graph denoised FBP (FBP-GD) 3) ART / SIRT reconstruction (without denosing) and 4) Graph denoised  ART / SIRT reconstruction (ART-GD, SIRT-GD).  Kaczmarz method with the relaxation parameter $\lambda = 0.25$ is used for ART and Cimmino's method is used for SIRT based reconstructions. Due to space constraints we only present results for Shepp-Logan and smooth phantoms. A close analysis of the results show that graph based denoising helps in attaining lower reconstruction error for analytical (FBP) as well as iterative (ART and SIRT) methods. This result is also visually obvious from the reconstructed phantoms. The error curves and intensity profiles (\ref{fig:results}) show that for both phantoms a lower error can be achieved by using graph denoised sinogram rather than regular raw data. Shepp-Logan reconstruction with ART specifically shows that the semi-convergence to an approximate solution is quick and divergence from this approximation is slower, an indication that the system has lower noise. The results show that the proposed method is extremely general and can be adapted for any tomographic reconstruction modality regardless of the reconstruction method employed.

 \begin{table}
  \centering
  \begin{threeparttable}[b]
  \captionsetup{justification=centering,   textfont={sc}}
  \caption{Comparison of Regular and Graph Denoised (GD) Reconstructions}
  \label{tab:test2}
  \begin{tabular}{lllll}
  \toprule
  \multicolumn{1}{c}{Phantom} & {FBP} & {FBP-GD\tnote{*}} & {ART\tnote{1}} & {ART-GD\tnote{*}}\\
   \cmidrule(l){1-5}
   Shepp-Logan (RN=0.05)        & 6.41       & 6.36       & 4.58   & 3.89  \\
   Shepp-Logan (RN=0.08)        & 6.76       & 6.53       & 5.53   & 4.44   \\
   \cmidrule(l){1-5}
  \multicolumn{1}{c}{Phantom} & {FBP} & {FBP-GD\tnote{*}} & {SIRT\tnote{2}} & {SIRT-GD\tnote{*}}\\
   \midrule
   Smooth (RN=0.05)        & 8.51       &  3.68      & 4.82   & 2.81  \\
   Smooth (RN=0.08)        & 13.16       & 4.83      & 6.65   & 3.82   \\
   \hline
  \end{tabular}
  \begin{tablenotes}
    \item[*] Proposed Method
  \end{tablenotes}
 \end{threeparttable}
\end{table}


%
%
%
%



%




\bibliographystyle{IEEEtran}
\bibliography{bibfile.bib}

\end{document}